\documentclass[10pt,conference]{IEEEtran}
\usepackage{cmap}
\usepackage{indentfirst}
\usepackage{booktabs}
\usepackage{enumitem,amsmath,amssymb}
\usepackage{tikz,pgfplots}  
\usepackage{cite}
\usepgfplotslibrary{colorbrewer}
\usetikzlibrary{shapes,arrows,fit,backgrounds,positioning,calc,matrix}
\pgfplotsset{compat=1.15}

\tikzset{
    /qrr/circle arrow/.cd,
    start angle/.initial={},
    delta angle/.initial={},
    end angle/.initial={},
    arrows/.estore in=\qrr@ca@arrow,
    arrows=-
}
\pgfdeclareshape{circle arrow}{
    \inheritsavedanchors[from=circle] \inheritanchorborder[from=circle]
    \inheritanchor[from=circle]{north}      \inheritanchor[from=circle]{north west}
    \inheritanchor[from=circle]{north east} \inheritanchor[from=circle]{center}
    \inheritanchor[from=circle]{west}       \inheritanchor[from=circle]{east}
    \inheritanchor[from=circle]{mid}        \inheritanchor[from=circle]{mid west}
    \inheritanchor[from=circle]{mid east}   \inheritanchor[from=circle]{base}
    \inheritanchor[from=circle]{base west}  \inheritanchor[from=circle]{base east}
    \inheritanchor[from=circle]{south}      \inheritanchor[from=circle]{south west}
    \inheritanchor[from=circle]{south east}
    \backgroundpath{
        \pgfkeysgetvalue{/qrr/circle arrow/start angle}\qrr@ca@s
        \pgfkeysgetvalue{/qrr/circle arrow/end angle}\qrr@ca@e
        \pgfkeysgetvalue{/qrr/circle arrow/delta angle}\qrr@ca@d
        \ifx\qrr@ca@s\pgfutil@empty%
            \pgfmathsetmacro\qrr@ca@s{\qrr@ca@e-\qrr@ca@d}%
        \else
            \ifx\qrr@ca@e\pgfutil@empty%
                \pgfmathsetmacro\qrr@ca@e{\qrr@ca@s+\qrr@ca@d}%
            \fi%
        \fi
        \pgfpathmoveto{\pgfpointadd{\centerpoint}{\pgfpointpolar{\qrr@ca@s}{\radius}}}%
        \pgfpatharc{\qrr@ca@s}{\qrr@ca@e}{\radius}%
        \pgfkeys{/tikz/arrows/.expand once=\qrr@ca@arrow}%
    }
}

\tikzset{
    turn left/.style={/tikz/shape=circle arrow,/qrr/circle arrow/arrows=->,/qrr/circle arrow/delta angle=340},
    turn right/.style={/tikz/shape=circle arrow,/qrr/circle arrow/arrows=<-,/qrr/circle arrow/delta angle=340},
    turn left north/.style  = {/tikz/turn left,  /qrr/circle arrow/start angle=100} ,
    turn left east/.style   = {/tikz/turn left,  /qrr/circle arrow/start angle=10},
    turn left south/.style  = {/tikz/turn left,  /qrr/circle arrow/start angle=280},
    turn left west/.style   = {/tikz/turn left,  /qrr/circle arrow/start angle=190},
    turn right north/.style = {/tikz/turn right, /qrr/circle arrow/start angle=100} ,
    turn right east/.style  = {/tikz/turn right, /qrr/circle arrow/start angle=10},
    turn right south/.style = {/tikz/turn right, /qrr/circle arrow/start angle=280},
    turn right west/.style  = {/tikz/turn right, /qrr/circle arrow/start angle=190},
}
\usepackage{lmodern}
\usepackage[T1]{fontenc}		
\usepackage[utf8]{inputenc}		
\usepackage{indentfirst}
\usepackage{adjustbox}
\usepackage{mathtools}
\usepackage{bm}
\usepackage{array}
\usepackage{multicol}
\usepackage{multirow}
\usepackage{lscape}
\usepackage{dblfloatfix}
\usepackage{float}
\usepackage{url}
\usepackage{graphicx}
\usepackage[linesnumbered,ruled]{algorithm2e}
\graphicspath{{figs/}}
\ifCLASSOPTIONcompsoc
 \usepackage[caption=false,font=normalsize,labelfont=sf,textfont=footnotesize]{subfig}
\else
 \usepackage[caption=false,font=footnotesize]{subfig}
\fi

\begin{document}
\title{FEMa-FS: Finite Element Machines for Feature Selection}
\author{
    \IEEEauthorblockN{Lucas Biaggi, João P. Papa,\\ Kelton A. P Costa}
    \IEEEauthorblockA{São Paulo State University \\
    Bauru, Brazil \\
    \{lucas.biaggi,joao.papa,\\kelton.costa\}@unesp.br} 
    \and 
    \IEEEauthorblockN{Danillo R. Pereira}
    \IEEEauthorblockA{Analytics2Go\\
    Álvares Machado, Brazil \\
    danillorobertopereirasds@gmail.com 
    }
    \and
    \IEEEauthorblockN{Leandro A. Passos}
    \IEEEauthorblockA{University of Wolverhampton\\ Wolverhampton, UK\\
    L.PassosJunior@wlv.ac.uk}
   }
\maketitle

\begin{abstract}
Identifying anomalies has become one of the primary strategies towards security and protection procedures in computer networks. 
In this context, machine learning-based methods emerge as an elegant solution to identify such scenarios and learn irrelevant information so that a reduction in the identification time and possible gain in accuracy can be obtained. This paper proposes a novel feature selection approach called Finite Element Machines for Feature Selection (FEMa-FS), which uses the framework of finite elements to identify the most relevant information from a given dataset. Although FEMa-FS can be applied to any application domain, it has been evaluated in the context of anomaly detection in computer networks. The outcomes over two datasets showed promising results.

\end{abstract}

\begin{IEEEkeywords}
Machine Learning, Feature Selection, Computer Networks Security, Finite Element Method
\end{IEEEkeywords}

\section{Introduction}
\label{s.intro}

The risk of anomalous activities in a computer network becomes one of the main concerns for security professionals, for they are in charge of identifying such activities and recognizing malicious attempts of unauthorized or illegal access~\cite{gargFuzzifiedCuckooBased2018}. Despite such professionals' efforts, the problem requires instantaneous response due to their unpredictable consequences, which attracted the attention of many researchers towards the development of intelligent and autonomous action plans~\cite{matelOptimizationNetworkIntrusion2019}. 

Machine learning strategies presented a considerable evolution in the last years. Among such techniques, one can refer to Finite Element Machines (FEMa)~\cite{pereiraFEMaFiniteElement2020}, which is a parameterless approach (under some circunstances) based on a numerical method analysis to find approximate solutions, i.e., the so-called Finite Element Method (FEM)~\cite{zienkiewiczFiniteElementMethod2013}. FEMa partitions the manifold that models the training data into simpler equations using basis functions by interpolating the dataset points. Later on, a version for regression purposes, called FEMaR (Finite Element Machines for Regression), was proposed by Pereira et al.~\cite{DanilloIJCNN:17} and further employed for reliability estimation of downhole safety valves~\cite{COLOMBO2020106894}.

Despite the advances mentioned above, machine learning approaches still face some challenges inherent to network security due to its intrinsic dynamics, which become more and more complex as computer systems evolve~\cite{falcaoQuantitativeComparisonUnsupervised2019}. Among some alternatives designed to alleviate the burden resulting from such complexities, one can refer to feature selection techniques, which can extract the most relevant information from data and discard redundant or irrelevant ones. Usually, such methods yield a more compact and representative dataset, later employed to feed some ML algorithms. It is expected that this compact representation provides more efficient and assertive classification to some extent~\cite{farisIntelligentSystemSpam2019}.

In this context, many works addressed the problem of feature selection using metaheuristic optimization techniques. Rodrigues et al~\cite{rodrigues2015binary}, for instance, proposed a binary version of the Flower Pollination Algorithm~\cite{yang2012flower, rodrigues2020adaptive} for feature selection, while Pereira et al.\cite{pereira2019jade} offered a similar approach using JADE~\cite{zhang2009jade}. The authors employed a graph-based Optimum-Path Forest classifier~\cite{PapaIJIST:2009,PapaPR:12} to evaluate the proposed methods' performance in both works. Other works also obtained satisfactory results, such as analysis of variance (ANOVA)~\cite{jcp1010011} and the $\chi^2$ independent variable test~\cite{kasongo2020performance}.

Despite the promising results, most approaches still suffer from drawbacks related to their stochastic nature and the challenges of avoiding local optima. This paper proposes a novel feature selection approach that uses FEMa attributes to select the best set of features, hereinafter called FEMa-FS. The idea is to compute the distance from each feature to the manifold learned by FEMa and sort them according to their relevance during the classificaton process.

The main contributions of this paper are summarized below:

\begin{itemize}
    \item to propose FEMa-FS, a novel technique for feature selection based on Finite Element Machines;
    \item to provide a new approach for network anomaly detection using FEMa-FS and Optimum-Path Forest; and
    \item to foster the literature in the context of feature selection, Finite Element Method, and network anomaly detection.
\end{itemize}


\section{Related Works}
\label{s:works}

Gharaee and Hosseinvand~\cite{gharaeeNewFeatureSelection2016} proposed a Genetic Algorithm (GA)-based approach for feature selection together with Support Vector Machines (SVM) to detect anomalies in computer networks. The combination achieved an accuracy rate higher than $99\%$ over the KDD-CUP 99 dataset, with a false positive rate lower than $1\%$. The researchers also conducted a performance evaluation over the UNSW-NB15 dataset with a broad variation in the accuracy rate. Even though, the false-positive rate stood below $0.10\%$. Feature selection approaches allowed the authors to find the features that best described different anomaly classes.

Khammassi and Krichen~\cite{khammassiGALRWrapperApproach2017} also introduced a feature selection approach based on GA to select the most relevant information for computer network anomaly detection. The approach worked together with linear regression to assess the relevance of each feature. Experiments conducted over KDD-CUP 99 and UNSW-NB15 datasets compared the proposed approach against three classification techniques, i.e., C4.5, Random Forest, and Naïve Bayes. The authors considered different scenarios concerning the number of features selected, ranging from $16$ to $18$, over KDD-CUP 99, and $18$ to $24$ considering UNSW-NB15 dataset. Besides, three distinct configurations regarding the number of training samples, i.e., $1,000$, $1,500$, and $2,000$, were employed in the experimental section.

In a similar work, Gottwalt et al.~\cite{gottwaltCorrCorrFeatureSelection2019} proposed the CorrCorr, a feature selection method for multivariate correlation-based network anomaly detection systems. The method outperformed the Principal Component Analysis and the Pearson class label correlation acknowledging the task of feature selection for network anomaly detection considering UNSWNB15 and NSL-KDD datasets.

Finally, Chkirbene et al.~\cite{chkirbeneTIDCSDynamicIntrusion2020} developed a framework to detect anomalies in computer networks composed of two models, i.e., the Trust-based Intrusion Detection and Classification System (TIDCS) and the Trust-based Intrusion Detection and Classification System- Accelerated (TIDCS-A). TIDCS randomly groups the features into clusters to further rank them according to their relevance and select the most representative ones. For evaluation purposes, the authors considered the NSL-KDD and the UNSW-NB15 datasets. The experiments presented satisfactory results, providing higher accuracy and lower false alarm rates than some state-of-the-art techniques. 

\section{Finite Element Machines}
\label{s.fema}

Let ${\cal D}=\{(\textbf{x}_i,y_i)\}_{i=1}^z$ be a dataset composed of training and testing partitions ${\cal D}_1$ and ${\cal D}_2$, respectively, where $\textbf{x}_i\in\Re^n$ stands for a given sample and $y_i\in\mathbb{N}$ its respective label. FEMa aims at learning a probabilistic manifold, i.e., a set of probability functions ${\cal F}(\textbf{x})=\{F_1(\textbf{x}),F_2(\textbf{x}),\ldots,F_c(\textbf{x})\}$ such that $F_j(\textbf{x})$ stands for the probability of $\textbf{x}\in{\cal D}_1$ be assigned to class $j$, and $c$ denotes the number of labels.

To explain the FEMa working mechanism, we started with the basic concepts about interpolating functions for further explaining how to use them to learn a manifold that encodes the entire training set.

\subsection{Interpolating Basis}
\label{ss.interpolating}

\begin{figure}[htb]
    \begin{tikzpicture}
        \begin{axis}[
              cycle list name=exotic, 
              legend style = {font =\footnotesize},
              legend pos = north west
        ]
        \addplot[
            only marks ] table [only marks,x=sample, y=value, col sep=comma] {samples.csv};
        \addplot+[
            smooth,
            thick,
            mark=none] table [x=sample, y=value, col sep=comma] {idw1k.csv};

        \addplot+[
            smooth,
            thick,
            mark=none] table [x=sample, y=value, col sep=comma] {idw2k.csv};
        \addplot+[
            smooth,
            thick,
            mark=none] table [x=sample, y=value, col sep=comma] {idw4k.csv};
        \addplot+[
            smooth,
            thick,
            mark=none] table [x=sample, y=value, col sep=comma] {idw6k.csv};
        \addplot+[
            smooth,
            thick,
            mark=none] table [x=sample, y=value, col sep=comma] {idw12k.csv};

        \addplot+[
            smooth,
            thick,
            mark=none] table [x=sample, y=value, col sep=comma] {idw15k.csv};

            \legend{sample,$k=1$,$k=2$,$k=4$,$k=6$,$k=12$,$k=15$}; 
        \end{axis}
    \end{tikzpicture}
   \caption{Shepard function for different values of $k$. Each black dot defines a training sample, and the curves stand for the interpolation (manifold) learned by FEMa.}\label{fig:sheF_idwpath}
\end{figure}
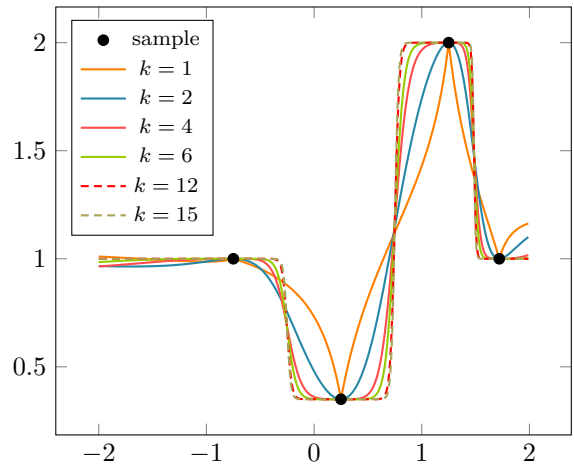

One of the most relevant advantages of FEMa concerns its light training step when we use bases functions that are interpolating natively~\cite{pereiraFEMaFiniteElement2020}. One common example is the Shepard basis~\cite{thackerAlgorithmXXXSHEPPACK2010}, which can be calculated as follows:

\begin{equation}
\label{eq.shepard_basis}
	\phi(\textbf{x},\textbf{x}_i;{\cal D}_1,k) = \frac{w(\textbf{x},\textbf{x}_i;k)}{\displaystyle\sum_{\textbf{x}_j\in{\cal D}_1}w(\textbf{x},\textbf{x}_j;k)},
\end{equation}
where $w$ is a non-negative function such that $w(\textbf{x},\textbf{x}_i)\rightarrow\infty$ when $\textbf{x}\rightarrow\textbf{x}_i$. In few words, if $\textbf{x}$ is closer to $\textbf{x}_i$, the function $w$ yields higher output values. Usually, the function $w$ employs a power $k\geq 1$ of the inverse of the Euclidean distance, as follows:

\begin{equation}
\label{e.w}
	w(\textbf{x},\textbf{x}_i;k) = \frac{1}{\lVert\textbf{x},\textbf{x}_i\rVert_2^k},
\end{equation}
where $\lVert\textbf{x},\textbf{x}_i\rVert_2$ denotes the Euclidean distance between $\textbf{x}$ and $\textbf{x}_i$. Moreover, $k$ is responsible for controlling the smoothness of the interpolation. Figure~\ref{fig:sheF_idwpath} shows different behaviors using Shepard bases with distinct values of $k$. Observe that higher values of $k$ generate sloppier curves.

Notice that non-interpolating basis, such as radial functions, requires using the inverse of a matrix to encode the expressiveness of each sample (basis element) and normalizing the basis function for cases in which it does not guarantee the ``partition-of-unity property". Even though FEMa is not restricted to specific functions, bases capable of guaranteeing interpolations and unit partitions are more appealing for processing large volumes of data, for they do not require computing the inverse matrix and the normalization as part of the training stage.

\subsection{Training Phase}
\label{ss.training}

Given a training sample $\textbf{x}\in{\cal D}_1$ and a basis function that obeys some properties~\cite{COLOMBO2020106894}, FEMa builds a probabilistic manifold that is an interpolation of bases functions $\phi(\cdot)$ centered at sample $\textbf{x}$:

\begin{equation}
\label{e.function_f}
	F_i(\textbf{x};{\cal D}_1,k)=\sum_{j=1}^{m}\rho_j^i\phi(\textbf{x},\textbf{x}_j;{\cal D}_1;k),\ i=1,2,\ldots,c,
\end{equation}
where $m$ denotes the number of training samples and $\rho_j^i\in[0,1]$ denotes the probability of training sample $\textbf{x}_j$ belonging to class $i$. One can observe that FEMa assigns a probability for each training sample considering all labels.

The probability $\rho_j^i$ is estimated as follows:

\begin{equation}
\label{e.probab}
\rho_j^i =
\begin{cases}
1 & \hbox{if $y_{j} = i$}\\  
0 & \hbox{otherwise.}\\
\end{cases}
\end{equation}

\subsection{Testing Phase}
\label{ss.testing}

Once training is finished, once can generate a set of probability functions ${\cal F}(\textbf{v})$ for each test sample $\textbf{v}\in{\cal D}_2$. The classification is performed as follows:

\begin{equation}
	\hat{y} = \arg\max_{i}F_i(\textbf{v}),
\end{equation}
where $\hat{y}$ is the label assigned to sample $\textbf{v}$.

FEMa also allows to infer the certainty in assigning the label $\hat{y}$ to each sample $\textbf{x}\in{\cal D}$ as follows:

\begin{equation}
\label{eq.certainty}
	C_{\hat{y}}(\textbf{x})=\frac{F_{\hat{y}}(\textbf{x})}{\sum_{i=1}^cF_i(\textbf{x})}.
\end{equation}
FEMa can deliver either a hard outcome or a soft classification, given by the degree of certainty.

\section{Feature selection through FEMa}
\label{s.FEMaFS}
 
This section presents the proposed approach, which adapts FEMa for feature selection purposes. Instead of associating a training sample $\bm{x} =[x^1,x^2,\ldots,x^n]$ to a given label, FEMa-FS learns how each feature $x^j$ from $\bm{x}$ contributes to its classification. FEMa-FS can be roughly divided into three primary steps: (i) feature normalization, (ii) probabilistic manifold learning, and (iii) computing the degree of feature overlap.

\subsection{Feature Normalization}
\label{ss.feature_normalization}

The first step aims at normalizing all dataset features as follows:

\begin{equation}
\label{e.feature_normalization}
    x_i^j = \frac{x_i^j-\min(x_l^j)}{\max(x_l^j)-\min(x_l^j)},\ l=1,2,\ldots,m,
\end{equation}
where $x_i^j\in[0,1]$ stands for the $j$-th feature of the $i$-th sample.

\subsection{Probabilistic Manifold Learning}
\label{ss.probabilistic_manifold_learning}

The following step concerns learn a manifold for each feature $i$ and class $c$. Let ${\cal Q} = \{q_1,q_2,\ldots,q_{p}\}$ be an ordered set of $p$ values such that $q_t\in[0,1]$. One can create a function $P_j^i(q_t;{\cal D}_1,k)$ that represents the probability of feature $j$ sampled at point $q_t$ belongs to class $i$ considering all training samples in ${\cal D}_1$ as follows: 

\begin{equation}
\label{e.probabilistic_function}
    P_j^i(q_t;{\cal D}_1,k) = \sum_{\bm{x}_l\in{\cal D}_1}\rho_l^i\phi_j(q_t,x_l^j;{\cal D}_1,k),
\end{equation}
where $\rho_l^i$ figures the same definition introduced in Equation~\ref{e.probab}, i.e., it stands for the probability of sample $\bm{x}_l$ belonging to class $i$. We assume that all features are normalized through Equation~\ref{e.feature_normalization} and the values in ${\cal Q}$ cover the domain $[0,1]$ sufficiently\footnote{Function $P_j^i$ has a similar role than function $F_i$ introduced in Equation~\ref{e.function_f}, but we decided to change the notation.}.

Let $\phi_j(\cdot)$ be a finite element basis function considering feature $j$ only, which can be computed as follows: 

\begin{equation}
\label{e.bsis_function_feature_j}
    \phi_j(q_t,x_l^j;{\cal D}_1,k) = \frac{w(q_t,x_l^j;k)}{\displaystyle\sum_{\bm{x}_k\in{\cal D}_1}w(q_t,x_k^j;k)},
\end{equation}
where the function $w$ has the same role as in Equation~\ref{e.w}, but here the parameters are single real numbers instead of an array. Once again, $k$ is in charge of controlling the smoothness of the interpolation process.

The idea is to learn a probabilistic manifold for each feature $j$, i.e., we compute a set of probabilistic functions $P_j=\{P_j^1,P_j^2,\ldots,P_j^c\}$ that represent the probability of assigning a given sample to one of the $c$ classes considering feature $j$ only\footnote{For the sake of clarity, we omitted the parameters of functions $P_j^i$ (Equation~\ref{e.probabilistic_function}).}.

\begin{figure}[!htb]
\centerline{
\begin{tabular}{c}
\includegraphics[width=6.57cm]{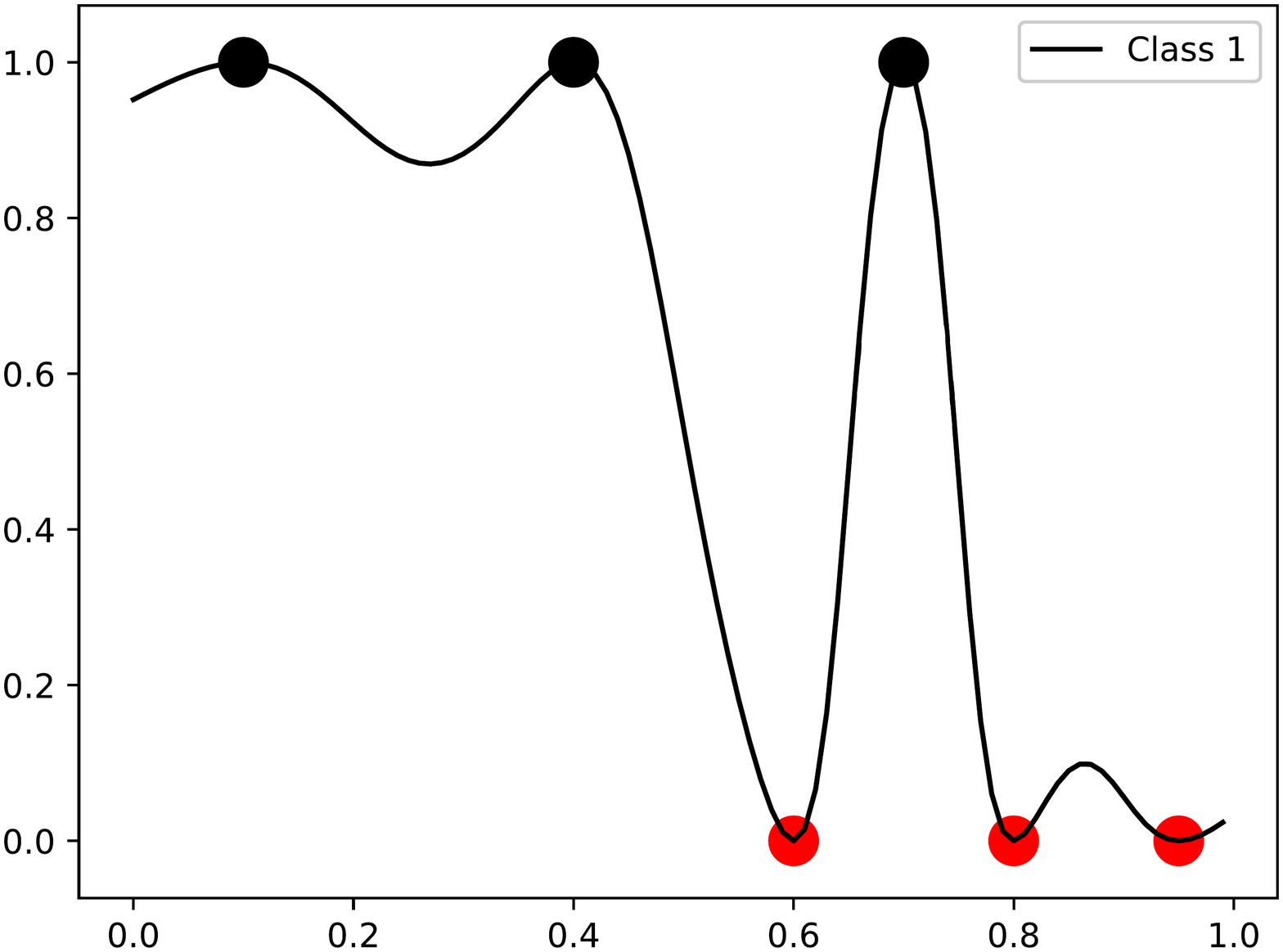} \\
(a)\\
\includegraphics[width=6.77cm]{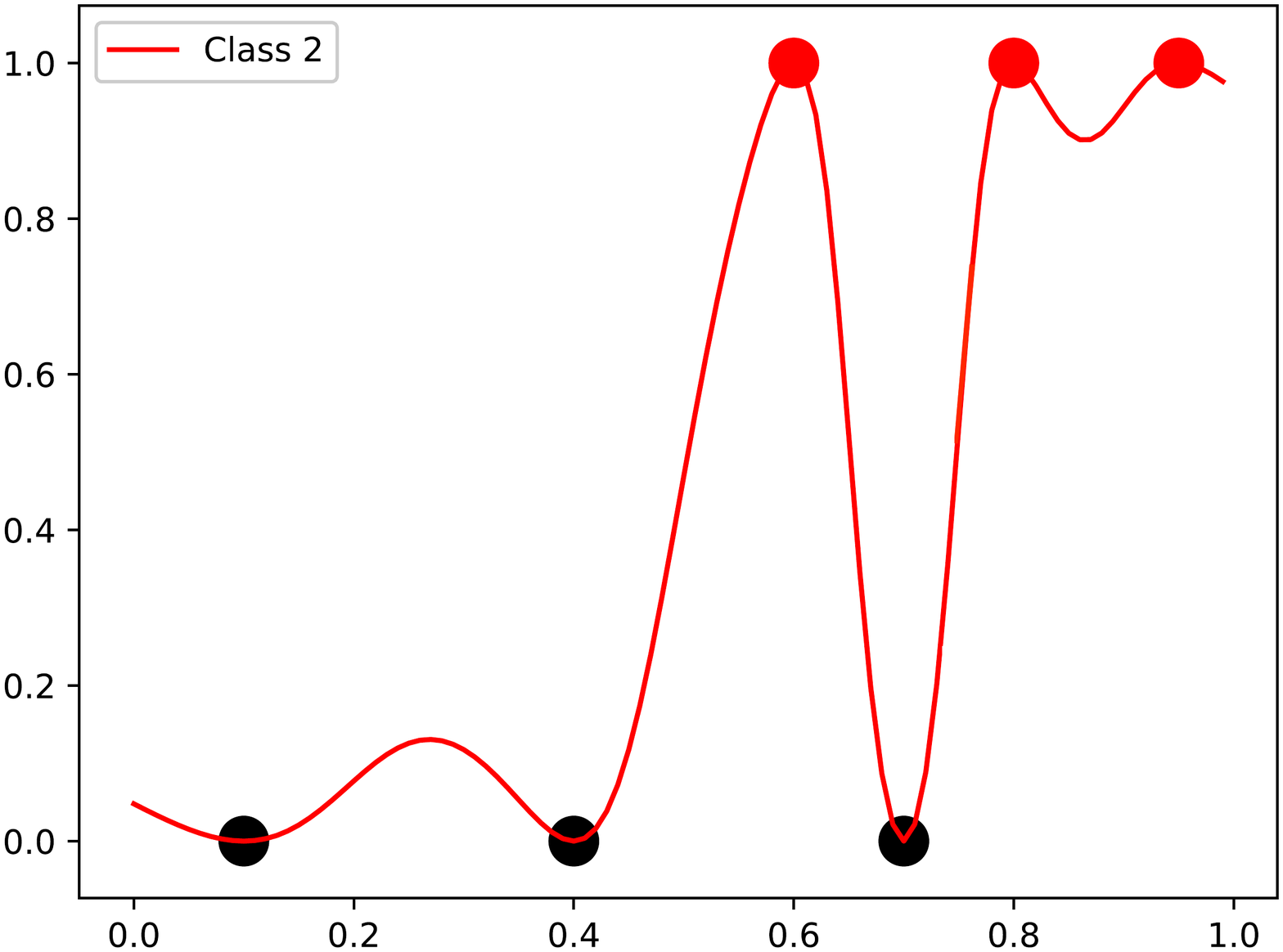} \\
(b)
\end{tabular}}	
\caption{Shepard approximation of the probability function of a two-class example using $k = 2$: (a) probability function regarding class $1$ - $P_j^1$ and (b) probability function regarding class $2$ - $P_j^2$. The red and black dots correspond to samples from classes $1$ and $2$, respectively.\label{f.probabilistic_functions}}
\end{figure}

Figure~\ref{f.probabilistic_functions} depicts a set of probabilistic functions learned by FEMa-FS considering a problem with $c=2$ classes and a given feature $j$. One can observe in Figure~\ref{f.probabilistic_functions}a that samples from class $1$, i.e., black dots, figure higher probabilities considering the black curve (probability function from class $1$ - $P_j^1$) than red dots. The opposite situation can be observed in Figure~\ref{f.probabilistic_functions}b.

\subsection{Computing the Degree of Feature Overlap}
\label{ss.feature_overlap}

\begin{figure}[!htb]
\centerline{
\begin{tabular}{c}
\includegraphics[width=6.17cm]{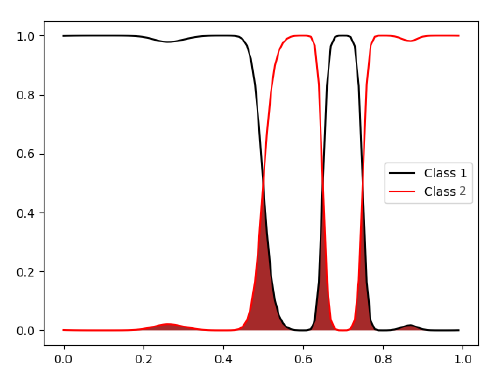} \\
(a)\\
\includegraphics[width=6.37cm]{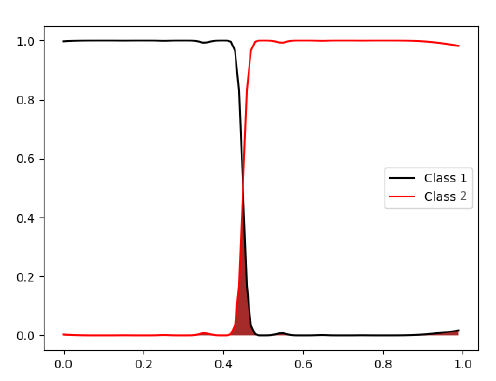} \\
(b)
\end{tabular}}	
\caption{Degree of overlap between two probability functions: (a) a ``bad"\ feature and (b) a ``good"\ feature.\label{f.overlap}}
\end{figure}

Last but not least, we need quantitative information to measure the quality of each feature. We propose to compute the average of the lowest probability between each class pair (for each feature), which is the overlap area between two probability functions. Therefore, the lower such a value, the better the quality of the feature. Figure~\ref{f.overlap} depicts such a situation.

The rationale is relatively straightforward, i.e., the lower the intersection area between two probability functions, the smaller the probability of confusion when deciding for one class or the other (considering a given feature). The number of selected features is an ad-hoc parameter that the user shall set.

\section{Methodology}
\label{s.methodology}

 
\subsection{Datasets}
\label{ss.dataset}

The UNSW-NB15 dataset~\cite{unswnb15} consists of two types of samples, i.e., normal network operations and synthetic generated anomalies. It comprises ten classes, one representing normal operations and nine denoting the following anomalies: Fuzzers, Analysis (e.g., port scans, email spams, and HTML files), Backdoor, DoS, Exploit, Generic, Reconnaissance, Shellcode, and Worm. In this work, we consider a binary classification approach, with standard network operations labeled as normal samples and synthetically generated anomalies are labeled as anomalies, i.e., all instances labeled as one of the nine anomaly types are gathered into a single class called ``anomaly". The attacks were conducted against different servers at the beginning of 2015 during $31$ hours. The dataset is divided into training and testing sets, such that the former comprises $175,341$ samples, while the latter is composed of  $82,332$ records, including all forms of attacks and usual traffic recordings. A preprocessing step was performed to keep the dataset homogeneity, i.e., the dataset was normalized to a standardized distribution in the interval $[0,1]$.

The second dataset, i.e., ISCXTor2016~\cite{habibi17}, consists of three users created for browser traffic collection and two for the communication parts such as chat, mail, FTP, and p2p. The traffic captured uses Wireshark and tcpdump, generating 22GB of data. To facilitate the labeling process, the outgoing traffic at the workstation and the gateway was simultaneous,  i.e., collecting a set of pairs of .pcap files: one for regular traffic and the other for Tor traffic file.

\subsection{Experimental Setup}
\label{ss.setup}

In this work, we consider Sheppard~\cite{shepardTwodimensionalInterpolationFunction1968} as the basis function and the Euclidean distance to calculate the distance between the feature values (function $w$ in Equation~\ref{e.bsis_function_feature_j}). The algorithm was compared against three baselines, i.e., $\chi^2$, ANOVA, and a plain classification approach with no feature selection, namely ``Baseline". The Optimum Path Forest classifier was used for classification purposes, for it is parameterless and has been used before on previous works successfully~\cite{rodrigues2015binary, pereira2019jade}. The techniques were compared using F1-score and accuracy measures.

Regarding the number of selected features, we considered eleven distinct scenarios, i.e., $10, 15, 20, 25, 30, 35, 40, 45, 50, 55$ and $60\%$ of the features. Each experiment was repeated during $25$ trials for statistical analysis using the Wilcoxon signed-rank test with $5\%$ of significance. FEMa-FS\footnote{Available at: \url{https://github.com/lbiaggi/femafs}} and OPF\footnote{Available at: \url{https://github.com/jppbsi/LibOPF}} were implemented using C language, while $\chi^2$ and ANOVA use sklearn implementations\footnote{Available at: \url{https://scikit-learn.org}}. The experiments were conducted using GNU/Arch Linux (64 bits) system with an Intel i7-3770K 3.50GHz processor and 24 GB of DDR3 RAM clocked at 1600MHz.
\section{Results}
\label{s.results}

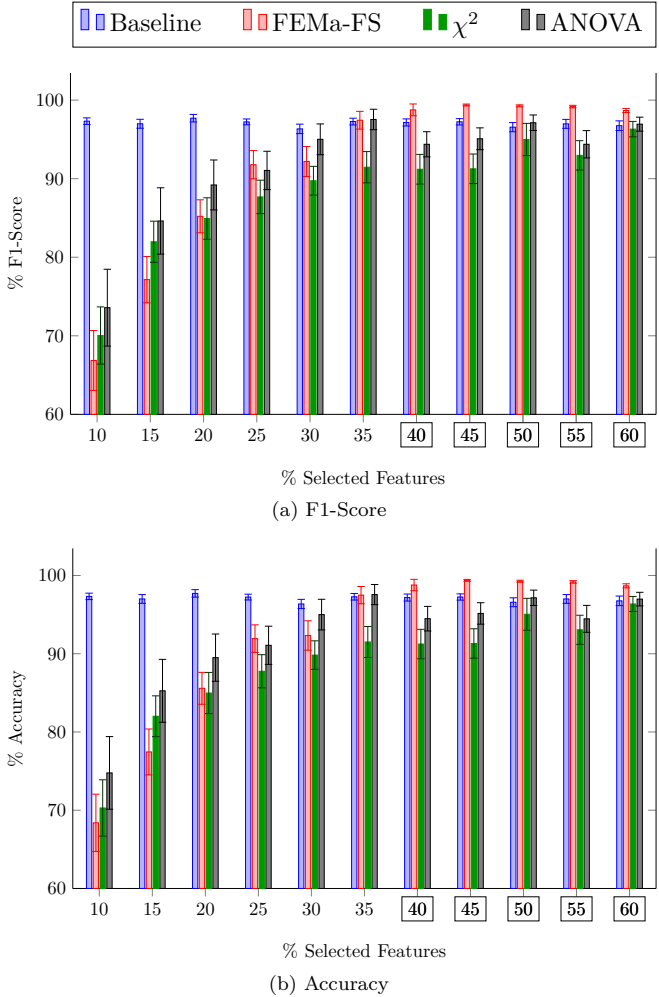
\begin{figure}[htb]
    \centering
    \begin{adjustbox}{width=0.48\textwidth}
         \hspace*{0.11\linewidth}\ref{columnsGraph}
    \end{adjustbox}
    \begin{adjustbox}{width=0.48\textwidth}
    \subfloat[][F1-Score]{\label{fig:shep_unswnb15_f}%
    \centering
    \begin{tikzpicture}[scale=0.75]
        \begin{axis}[width=0.49\textwidth, height = 8 cm,symbolic x coords={10,15,20,25,30,35,40,45,50,55,60},  bar width=0.10cm,
        ymin=60,x=1cm,xmin=10, xmax=60, axis x line=bottom,
            ylabel={ \% F1-Score },
            xlabel={ \% Selected Features},
            enlarge x limits=0.05, 
            ybar=2*\pgflinewidth,
            /pgf/number format/1000 sep={},
            extra x ticks = {40,45,50,55,60},
            extra x tick style={tick label style={draw=black}},
            legend columns=-1,
            legend style={/tikz/every even column/.append style={column sep=0.5cm}},
            axis line style={-},
            legend to name=columnsGraph,
            ]
            \addplot+ [ error bars/.cd, y dir=both, y explicit,] coordinates {
                (10,97.32) +- (10,0.42)
                (15,96.99) +- (15,0.57)
                (20,97.7) +- (20,0.48)
                (25,97.23) +- (25,0.37)
                (30,96.34) +- (30,0.6)
                (35,97.27) +- (35,0.43)
                (40,97.16) +- (40,0.45)
                (45,97.24) +- (45,0.4)
                (50,96.56) +- (50,0.58)
                (55,96.98) +- (55,0.57)
                (60,96.74) +- (60,0.62)
            };
            \addplot+ [error bars/.cd, y dir=both, y explicit,] coordinates {
                (10,66.85) +- (10,3.82)
                (15,77.15) +- (15,2.95)
                (20,85.21) +- (20,2.10)
                (25,91.78) +- (25,1.80)
                (30,92.18) +- (30,1.91)
                (35,97.44) +- (35,1.13)
                (40,98.76) +- (40,0.75)
                (45,99.38) +- (45,0.12)
                (50,99.27) +- (50,0.13)
                (55,99.18) +- (55,0.16)
                (60,98.68) +- (60,0.26)

            };
            
            \addplot+ [ color=green!60!black, error bars/.cd, y dir=both, y explicit, 
            error bar style={color=black!80!green}] coordinates {
              (10,70.04) +- (10,3.64)
              (15,81.97) +- (15,2.62)
              (20,84.93) +- (20,2.64)
              (25,87.68) +- (25,2.13)
              (30,89.74) +- (30,1.83)
              (35,91.46) +- (35,1.99)
              (40,91.20) +- (40,1.89)
              (45,91.26) +- (45,1.88)
              (50,94.99) +- (50,2.05)
              (55,92.97) +- (55,1.87)
              (60,96.31) +-  (60,0.98)
            };
            \addplot+ [ error bars/.cd, y dir=both, y explicit, ] coordinates {
                (10,73.58) +- (10,4.89)
                (15,84.62) +- (15,4.22)
                (20,89.20) +- (20,3.18)
                (25,91.05) +- (25,2.44)
                (30,95.01) +- (30,1.96)
                (35,97.54) +- (35,1.30)
                (40,94.38) +- (40,1.60)
                (45,95.08) +- (45,1.39)
                (50,97.12) +- (50,0.99)
                (55,94.38) +- (55,1.74)
                (60,96.94) +- (60,0.89)
            };
            \legend{Baseline, FEMa-FS, $\chi^2$, ANOVA}; 
        \end{axis}
    \end{tikzpicture}%
    }
    \end{adjustbox}
    \begin{adjustbox}{width=0.48\textwidth}
    \subfloat[][Accuracy]{\label{fig:shep_unswnb15_a}%
    \begin{tikzpicture}[scale=0.75]
        \begin{axis}[width=0.49\textwidth, height = 8 cm,symbolic x coords={10,15,20,25,30,35,40,45,50,55,60},  bar width=0.10cm, x=1cm,xmin=10, xmax=60, axis x line=bottom,
            enlarge x limits=0.05, 
            ybar=2*\pgflinewidth,
            ymin=60,
            ylabel={ \% Accuracy },
            xlabel={ \% Selected Features },
            /pgf/number format/1000 sep={},
            extra x ticks = {40,45,50,55,60},
            extra x tick style={tick label style={draw=black}},
            legend columns=-1,
            legend style={/tikz/every even column/.append style={column sep=0.25cm}},
            axis line style={-}
            ]
            \addplot+ [ error bars/.cd, y dir=both, y explicit, ] coordinates {

                (10,97.33) +- (10,0.41)
                (15,97.00) +- (15,0.56)
                (20,97.71) +- (20,0.48)
                (25,97.24) +- (25,0.37)
                (30,96.36) +- (30,0.59)
                (35,97.28) +- (35,0.42)
                (40,97.18) +- (40,0.45)
                (45,97.25) +- (45,0.4)
                (50,96.58) +- (50,0.57)
                (55,97.00) +- (55,0.56)
                (60,96.76) +- (60,0.61)

            };
            \addplot+ [error bars/.cd, y dir=both, y explicit,] coordinates {

                (10,68.39) +- (10,3.66)
                (15,77.45) +- (15,2.93)
                (20,85.58) +- (20,2.04)
                (25,91.93) +- (25,1.76)
                (30,92.33) +- (30,1.88)
                (35,97.49) +- (35,1.1)
                (40,98.78) +- (40,0.73)
                (45,99.38) +- (45,0.12)
                (50,99.27) +- (50,0.13)
                (55,99.18) +- (55,0.16)
                (60,98.68) +- (60,0.26)

            };

            \addplot+ [ color=green!60!black, error bars/.cd, y dir=both, y explicit, 
            error bar style={color=black!80!green}]  coordinates {

                (10,70.29) +- (10,3.6)
                (15,82.01) +- (15,2.61)
                (20,84.98) +- (20,2.63)
                (25,87.76) +- (25,2.12)
                (30,89.83) +- (30,1.82)
                (35,91.50) +- (35,1.98)
                (40,91.24) +- (40,1.88)
                (45,91.31) +- (45,1.88)
                (50,95.04) +- (50,2.04)
                (55,93.06) +- (55,1.86)
                (60,96.36) +- (60,0.96)

            };
            \addplot+ [ error bars/.cd, y dir=both, y explicit, ] coordinates {
                (10,74.77) +- (10,4.64)
                (15,85.26) +- (15,4.02)
                (20,89.5) +- (20,3.02)
                (25,91.08) +- (25,2.44)
                (30,95.00) +- (30,1.96)
                (35,97.56) +- (35,1.29)
                (40,94.48) +- (40,1.57)
                (45,95.15) +- (45,1.37)
                (50,97.15) +- (50,0.98)
                (55,94.46) +- (55,1.72)
                (60,96.98) +- (60,0.87)

            };
        \end{axis}
    \end{tikzpicture}%
    }
    \end{adjustbox}
\caption{Experimental results obtained over UNSW-NB15 dataset: (a) F1-score and (b) accuracy. We considering $11$ scenarios with different percentages of selected features.}
\label{fig:femafs_shep_unswnb15_graphics}
\end{figure}

This section presents the experimental results to evaluate the robustness of the proposed approach. Figure~\ref{fig:femafs_shep_unswnb15_graphics} presents the F1-Score and accuracy results concerning anomaly detection in the UNSW-NB15 dataset. The squared values in the $x$-label stand for the scenarios where FEMa-FS obtained statistically superior results to the other approaches. The best results concerning F1-Score and accuracy were obtained using $45\%$ of the features only, in which FEMa-FS overpassed the baseline by $2.14\%$.

For the sake of visualization purposes, Figure~\ref{fig:femafs_shep_unswnb15} shows the confusion matrix regarding the baseline and the scenario composed of $45\%$ of features selected by FEMa-FS, i.e., the configuration that obtained the best results. One can observe that the true positives (FP) and true negatives (TN) are similar between both approaches. On the other hand, FEMa-FS obtained fewer misclassification results, i.e., the number of false positives (FP) and false negatives (FN) are considerably smaller than the baseline. In the context of intrusion detection in computer networks, false negatives are particularly important, for they stand for attacks that were not detected by the intrusion detection system.

\begin{figure}[!h]
    \centering
    \subfloat[][Baseline]{\label{fig:CM_usnwb15_opf}%
    \centering
    \begin{tikzpicture}[
            scale=1.25
        ] 
        \draw[thick, color=blue] (0,0) -- (2.2,0);
        \draw[thick, color=blue] (0,0) -- (0, 2.2);
        \draw[thick, color=blue] (2.2,2.2) -- (2.2, 0);
        \draw[thick, color=blue] (2.2,2.2) -- (0, 2.2);

        \draw[thick, color=blue] (-0.3, 1.1) -- (2.2, 1.1);
        \draw[thick, color=blue] (1.1, 0) -- (1.1, 2.5);

        \coordinate[label=left:($+$)] (p1) at (-0.1,1.6);
        \coordinate[label=left:($-$)] (p2) at (-0.1,0.4);

        \coordinate[label=above:($+$)] (p3) at (0.55, 2.2);
        \coordinate[label=above:($-$)] (p4) at (1.65, 2.2);


        \coordinate[label={ \scriptsize TP$\,=4501$}, color=blue] (TP) at (0.55, 1.50);
        \coordinate[label={ \scriptsize FP$\,=189$}, color=blue] (FP) at (1.65, 1.50);
        \coordinate[label={ \scriptsize FN$\,=37$}, color=blue] (FN) at (0.55, 0.40);
        \coordinate[label={ \scriptsize TN$\,=3506$}, color=blue] (TN) at (1.65, 0.40);
    \end{tikzpicture}
    }
    \qquad
    \subfloat[][FEMa-FS]{\label{fig:CM_usnwb15}%
    \centering
    \begin{tikzpicture}[
            scale=1.25
        ] 
        \draw[thick, color=red] (0,0) -- (2.2,0);
        \draw[thick, color=red] (0,0) -- (0, 2.2);
        \draw[thick, color=red] (2.2,2.2) -- (2.2, 0);
        \draw[thick, color=red] (2.2,2.2) -- (0, 2.2);

        \draw[thick, color=red] (-0.3, 1.1) -- (2.2, 1.1);
        \draw[thick, color=red] (1.1, 0) -- (1.1, 2.5);

        \coordinate[label=left:($+$)] (p1) at (-0.1,1.6);
        \coordinate[label=left:($-$)] (p2) at (-0.1,0.4);

        \coordinate[label=above:($+$)] (p3) at (0.55, 2.2);
        \coordinate[label=above:($-$)] (p4) at (1.65, 2.2);


        \coordinate[label={ \scriptsize TP$\,=4521$}] (TP) at (0.55, 1.50);
        \coordinate[label={ \scriptsize FP$\,=34$}] (FP) at (1.65, 1.50);
        \coordinate[label={ \scriptsize FN$\,=17$}] (FN) at (0.55, 0.40);
        \coordinate[label={ \scriptsize TN$\,=3661$}] (TN) at (1.65, 0.40);
    \end{tikzpicture}
    }
    \qquad
  \caption{Confusion matrices considering UNSW-NB15 dataset: (a) baseline (standard OPF) and (b) FEMa-FS with $45\%$ of features selected.}\label{fig:femafs_shep_unswnb15}
\end{figure}


Table~\ref{tab:femafs_shep_usnwb15_wx_f1} presents the statistical analysis between FEMa-FS and $\chi^2$, ANOVA, and the baseline considering the Wilcoxon signed-rank test with $5\%$ of significance over the F1-score results in the UNSW-NB15 dataset. The symbol $=$ denotes that the technique's results are similar to FEMa-FS, $\uparrow$ stands for the cases where the technique was more accurate than FEMa-FS, and $\downarrow$ represents the scenarios in which FEMa-FS outperformed the method statistically. Such results show that even though FEMa-FS did not perform as good as the other techniques considering scenarios from $10\%$ to $35\%$ of the features, the results are statistically similar in most cases, except for the baseline considering FEMa-FS with $10\%$, $15\%$, and $20\%$ the features, and ANOVA with $15\%$ and $35\%$ of the features. For the remaining scenarios, FEMa-FS performed equal or better than the other techniques, confirming the robustness of the proposed approach.


 Figure~\ref{fig:femafs_shep_icsxtor_graphics} provides the respective F1-score and accuracy outcomes regarding ISCXTor2016 dataset. In such context, one can observe that FEMa-FS obtained the best results overall, considering all scenarios with more than $30\%$ of the samples. Additionally, FEMa-FS performed statistically better than ANOVA and $\chi^2$ in all scenarios, obtaining results similar to or better than the baseline considering $20\%$ or more features, as presented in Table~\ref{tab:femafs_shep_ICSXTor2016_wx_f1}.

\begin{table}[hbt]
     \centering
     \caption{Statistical analysis of F1-score results compared to FEMa-FS considering the Wilcoxon signed-rank test with $5\%$ of significance over UNSW-NB15 dataset}.\label{tab:femafs_shep_usnwb15_wx_f1}
    \begin{adjustbox}{width=0.49\textwidth}
    \begin{tabular}{*{12}{c}}
       \cr\toprule
       Technique & 10\% & 15\% & 20\% & 25\% & 30\% & 35\% & 40\% & 45\% & 50\% & 55\% & 60\% \\\midrule
       $\chi^2$   & = & = & = & = & = & = & = & $\downarrow$ & = & = & $\downarrow$  \\\midrule 
       ANOVA     & = & $\uparrow$ & = & = & = & $\uparrow$ & = & = & = & = & = \\\midrule
       Baseline  & $\uparrow$ & $\uparrow$ & $\uparrow$ & = & = & $\downarrow$ & $\downarrow$ & $\downarrow$ & $\downarrow$ & $\downarrow$ & $\downarrow$ \\\bottomrule
    \end{tabular}
    \end{adjustbox}
\end{table} 
{\color{black}
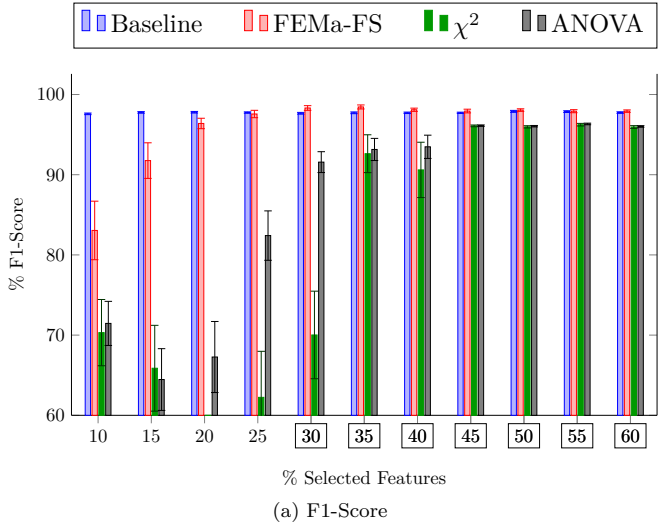
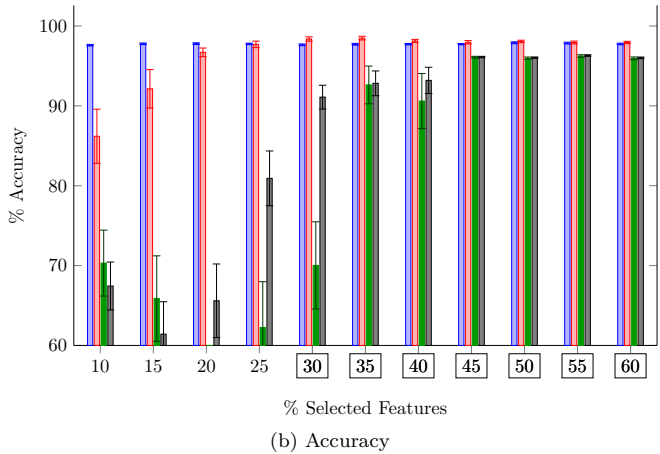
\begin{figure}[htb]
    \centering
    \begin{adjustbox}{width=0.48\textwidth}
         \hspace*{0.11\linewidth}\ref{columnsGraphTOR}
    \end{adjustbox}
    \begin{adjustbox}{width=0.48\textwidth}
    \subfloat[][F1-Score]{\label{fig:shep_icsxtor_f}%
    \centering
    \begin{tikzpicture}[scale=0.75]
        \begin{axis}[width=0.49\textwidth, height = 8 cm,symbolic x coords={10,15,20,25,30,35,40,45,50,55,60},  bar width=0.10cm,
        ymin=60,x=1cm,xmin=10, xmax=60, axis x line=bottom,
            ylabel={ \% F1-Score },
            xlabel={ \% Selected Features},
            enlarge x limits=0.05, 
            ybar=2*\pgflinewidth,
            /pgf/number format/1000 sep={},
            extra x ticks = {30,35,40,45,50,55,60},
            extra x tick style={tick label style={draw=black}},
            legend columns=-1,
            legend style={/tikz/every even column/.append style={column sep=0.5cm}},
            axis line style={-},
            legend to name=columnsGraphTOR,
            ]
                \addplot+ [ error bars/.cd, y dir=both, y explicit, ] coordinates {
                        (10,97.59) +- (10,0.1)
                        (15,97.77) +- (15,0.1)
                        (20,97.79) +- (20,0.1)
                        (25,97.75) +- (25,0.09)
                        (30,97.66) +- (30,0.11)
                        (35,97.72) +- (35,0.11)
                        (40,97.72) +- (40,0.09)
                        (45,97.72) +- (45,0.08)
                        (50,97.9) +- (50,0.12)
                        (55,97.87) +- (55,0.11)
                        (60,97.75) +- (60,0.1)

                    };

                    \addplot+ [error bars/.cd, y dir=both, y explicit,] coordinates {
                        (10,83.05) +- (10,3.65)
                        (15,91.76) +- (15,2.22)
                        (20,96.39) +- (20,0.65)
                        (25,97.57) +- (25,0.46)
                        (30,98.32) +- (30,0.29)
                        (35,98.45) +- (35,0.25)
                        (40,98.1) +- (40,0.2)
                        (45,97.94) +- (45,0.23)
                        (50,98.06) +- (50,0.16)
                        (55,97.92) +- (55,0.19)
                        (60,97.93) +- (60,0.15)
                    };

                    \addplot+ [ color=green!60!black, error bars/.cd, y dir=both, y explicit,
                    error bar style={color=black!80!green}] coordinates {
                        (10,70.31) +- (10,4.13)
                        (15,65.87) +- (15,5.35)
                        (20,47.72) +- (20,6.1)
                        (25,62.24) +- (25,5.74)
                        (30,70.02) +- (30,5.46)
                        (35,92.62) +- (35,2.36)
                        (40,90.6) +- (40,3.45)
                        (45,96.09) +- (45,0.11)
                        (50,95.97) +- (50,0.14)
                        (55,96.24) +- (55,0.16)
                        (60,95.95) +- (60,0.17)
                    };
                    \addplot+ [error bars/.cd, y dir=both, y explicit,] coordinates {
                        (10,71.46) +- (10,2.75)
                        (15,64.46) +- (15,3.85)
                        (20,67.27) +- (20,4.43)
                        (25,82.41) +- (25,3.08)
                        (30,91.57) +- (30,1.3)
                        (35,93.15) +- (35,1.38)
                        (40,93.48) +- (40,1.45)
                        (45,96.12) +- (45,0.09)
                        (50,96.05) +- (50,0.09)
                        (55,96.33) +- (55,0.1)
                        (60,96.02) +- (60,0.1)
                    };

            \legend{Baseline, FEMa-FS, $\chi^2$, ANOVA}; 
        \end{axis}
    \end{tikzpicture}%
    }
    \end{adjustbox}
    \begin{adjustbox}{width=0.48\textwidth}
    \subfloat[][Accuracy]{\label{fig:shep_icsxtor_a}%
    \begin{tikzpicture}[scale=0.75]
        \begin{axis}[width=0.49\textwidth, height = 8 cm,symbolic x coords={10,15,20,25,30,35,40,45,50,55,60},  bar width=0.10cm, x=1cm,xmin=10, xmax=60, axis x line=bottom,
            enlarge x limits=0.05, 
            ybar=2*\pgflinewidth,
            ymin=60,
            ylabel={ \% Accuracy },
            xlabel={ \% Selected Features },
            /pgf/number format/1000 sep={},
            extra x ticks = {30,35,40,45,50,55,60},
            extra x tick style={tick label style={draw=black}},
            legend columns=-1,
            legend style={/tikz/every even column/.append style={column sep=0.25cm}},
            axis line style={-}
            ]
            \addplot+ [ error bars/.cd, y dir=both, y explicit, ] coordinates {
                    (10,97.59) +- (10,0.1)
                    (15,97.77) +- (15,0.1)
                    (20,97.79) +- (20,0.1)
                    (25,97.76) +- (25,0.09)
                    (30,97.64) +- (30,0.11)
                    (35,97.71) +- (35,0.11)
                    (40,97.72) +- (40,0.09)
                    (45,97.72) +- (45,0.08)
                    (50,97.9) +- (50,0.12)
                    (55,97.86) +- (55,0.11)
                    (60,97.75) +- (60,0.1)
                };

                \addplot+ [error bars/.cd, y dir=both, y explicit,] coordinates {
                    (10,86.19) +- (10,3.39)
                    (15,92.13) +- (15,2.41)
                    (20,96.7) +- (20,0.54)
                    (25,97.7) +- (25,0.41)
                    (30,98.37) +- (30,0.27)
                    (35,98.48) +- (35,0.23)
                    (40,98.13) +- (40,0.19)
                    (45,97.96) +- (45,0.21)
                    (50,98.07) +- (50,0.15)
                    (55,97.93) +- (55,0.18)
                    (60,97.93) +- (60,0.15)

                };
                \addplot+ [ color=green!60!black, error bars/.cd, y dir=both, y explicit,
                error bar style={color=black!80!green}] coordinates {
                    (10,70.31) +- (10,4.13)
                    (15,65.87) +- (15,5.35)
                    (20,47.72) +- (20,6.1)
                    (25,62.24) +- (25,5.74)
                    (30,70.02) +- (30,5.46)
                    (35,92.62) +- (35,2.36)
                    (40,90.6) +- (40,3.45)
                    (45,96.09) +- (45,0.11)
                    (50,95.97) +- (50,0.14)
                    (55,96.24) +- (55,0.16)
                    (60,95.95) +- (60,0.17)
                };
                \addplot+ [error bars/.cd, y dir=both, y explicit,] coordinates {
                    (10,67.44) +- (10,3.0)
                    (15,61.41) +- (15,4.06)
                    (20,65.59) +- (20,4.61)
                    (25,80.92) +- (25,3.43)
                    (30,91.09) +- (30,1.49)
                    (35,92.82) +- (35,1.55)
                    (40,93.19) +- (40,1.64)
                    (45,96.11) +- (45,0.09)
                    (50,96.03) +- (50,0.09)
                    (55,96.30) +- (55,0.10)
                    (60,96.01) +- (60,0.10)
                };

        \end{axis}
    \end{tikzpicture}%
    }
    \end{adjustbox}
\caption{Experimental results obtained over  ICSXTor2016 dataset: (a) F1-score and (b) accuracy. We considering $11$ scenarios with different percentages of selected features.}
\label{fig:femafs_shep_icsxtor_graphics}
\end{figure}
\begin{table}[hbt]
     \centering
     \caption{Statistical analysis of F1-score results compared to FEMa-FS considering the Wilcoxon signed-rank test with $5\%$ of significance over ICSXTor2016 dataset}.\label{tab:femafs_shep_ICSXTor2016_wx_f1}
    \begin{adjustbox}{width=0.49\textwidth}
    \begin{tabular}{*{12}{c}}
       \cr\toprule
       Technique & 10\% & 15\% & 20\% & 25\% & 30\% & 35\% & 40\% & 45\% & 50\% & 55\% & 60\% \\\midrule
       $\chi^2$   &  $\downarrow$  &  $\downarrow$  &  $\downarrow$ &  $\downarrow$ &  $\downarrow$ & $\downarrow$ & $\downarrow$ & $\downarrow$ &  $\downarrow$ &  $\downarrow$ & $\downarrow$  \\\midrule 
       ANOVA     &  $\downarrow$ &  $\downarrow$ &  $\downarrow$ &  $\downarrow$ &  $\downarrow$ &  $\downarrow$ &  $\downarrow$ &  $\downarrow$ &  $\downarrow$ &  $\downarrow$ &  $\downarrow$ \\\midrule
       Baseline  & $\uparrow$ & $\uparrow$ & = & = &  $\downarrow$ & $\downarrow$ & $\downarrow$ & $\downarrow$ & = & = & = \\\bottomrule
    \end{tabular}
    \end{adjustbox}
\end{table} 
}

Table~\ref{tab:femafs_comp} presents a comparison of the proposed approach against other results from the literature considering the UNSW-NB15 dataset. One can observe that FEMa-FS obtained the best results considering the accuracy values. Regarding the True Positive Rate (TPR), FEMa-FS outperformed all techniques except~\cite{gottwaltCorrCorrFeatureSelection2019}. Similar results are observed considering the False Positive Rate (FPR), in which FEMa-FS outperformed all techniques except~\cite{gharaeeNewFeatureSelection2016}. Such results reinforce that FEMa-FS is a powerful tool for feature selection.

\begin{table}[hbt]
    \centering
    \caption{Comparison with works from the literature over UNSW-NB15 dataset}.\label{tab:femafs_comp}
 \begin{adjustbox}{width=0.49\textwidth}   
     \begin{tabular}{*{4}{c}}
        \cr\toprule
        Work & Accuracy & TPR & FPR \\\midrule
        Gharaee and Hosseinvand~\cite{gharaeeNewFeatureSelection2016} & 93.25\% & 90.75\% & \textbf{0.04}\% \\\midrule
        Khammassi and Krichen~\cite{khammassiGALRWrapperApproach2017} & 81.42\% & - & 6.39\% \\\midrule
        Gottwalt et al.~\cite{gottwaltCorrCorrFeatureSelection2019} & 93.22\% & \textbf{100\%} & 7.88\% \\\midrule
        Almomani~\cite{almomaniFeatureSelectionModel2020} & 90.48\% & 97.14\% & 14.96\% \\\midrule
        Chkirbene et al.~\cite{chkirbeneTIDCSDynamicIntrusion2020} & 91\% & 94\% & 4\% \\\midrule
        \textit{OPF (Baseline)} & 97.24\% & 99.18\% & 5.12\% \\\midrule
        \textbf{FEMa-FS (proposed)} & \textbf{99.38}\% & 99.63\% & 0.92\% \\\bottomrule
 \end{tabular}
 \end{adjustbox}
\end{table}

\section{Conclusions and Future Works}
\label{s.conclusion}

This work proposes FEMa-FS, a novel method for feature selection inspired by the Finite Element Machines classifier that considers interpolation basis functions to construct a probabilistic manifold to label the dataset samples. The importance of each feature is evaluated during the manifold learning process so that the top features are further used for the classification process. The proposed approach showed promising results, outperforming a standard classification with no feature selection (i.e., baseline) and other well-known approaches such as ANOVA and $\chi^2$, as well as other works from the literature. 

Regarding future works, we aim at extending FEMa-FS using different distances, such as Manhattan and Hamming, as well as to evaluate the performance considering other basis functions~\cite{thackerAlgorithmXXXSHEPPACK2010}. To reduce processing time, we shall implement an optimized version based on kd-trees.


\section*{Acknowledgments}
The authors are grateful to FAPESP grants \#2021/05516-1, \#2017/22905-6, \#2013/07375-0, \#2014/12236-1, and \#2016/19403-6, the Brazilian National Council for Research and Development (CNPq) via grants No. 429003/2018 —8, 304315/2017 —6, 430274/2018 —1, 307066/2017 —7 and 427968/2018 —6, as well as the Engineering and Physical Sciences Research Council (EPSRC) grant EP/T021063/1 and its principal investigator Ahsan Adeel.

\bibliographystyle{IEEEtran}
\bibliography{main_biber}

\end{document}